# Inducing Sparse Coding and And-Or Grammar from Generator Network

Xianglei Xing[1,2*], Song-Chun Zhu[2], Ying Nian Wu[2]
[1]College of Automation, Harbin Engineering University, Harbin 150001, China
[2]Department of statistics, University of California, Los Angeles, California 90095
xingxl@hrbeu.edu.cn, sczhu@stat.ucla.edu, ywu@stat.ucla.edu

## Abstract

We introduce an explainable generative model by applying sparse operation on the feature maps of the generator network. Meaningful hierarchical representations are obtained using the proposed generative model with sparse activations. The convolutional kernels from the bottom layer to the top layer of the generator network can learn primitives such as edges and colors, object parts, and whole objects layer by layer. From the perspective of the generator network, we propose a method for inducing both sparse coding and the AND-OR grammar for images. Experiments show that our method is capable of learning meaningful and explainable hierarchical representations.

## 1 Introduction

The sparsity principle has played a fundamental role in high-dimensional statistics, machine learning, and signal processing. In particular, sparse coding (Olshausen and Field 1996) is an important principle for understanding the visual cortex. By imposing sparsity constraints on the coefficients of a linear generative model, (Olshausen and Field 1997) learn Gabor-like wavelets that resemble the neurons in the primary visual cortex (V1) from natural image patches.

However, developing a top-down sparse coding model that can generate (in addition to merely reconstruct) realistic natural image patterns has proven to be a difficult task. The model of (Olshausen and Field 1997) assumes that the coefficients of the linear model follow a sparse distribution. However, it is difficult to develop a realistic model for the sophisticate sparse patterns, which involves a selection process that selects which coefficients are active, in addition to generating the values of the active coefficients. The commonly used independent spike and slab model (Ishwaran and Rao 2005) can hardly generate realistic images. Moreover, even if we do have such a selection model as a prior model for the coefficients, fitting the model to each training example involves a non-trivial inference process to identify the active coefficients and estimate their values. Because of the modeling and computing difficulties, we still do not have a realistic top-down generative model that has the linear sparse coding model at the lowest layer or incorporates the sparse coding principle across its layers.

Recently, the generator network (Goodfellow et al. 2014; Higgins et al. 2016; Xing et al. 2018) has proven to be a surprisingly powerful top-down model for natural images (as well as other types of signals). This model can also be considered a generalization of the factor analysis model. While the sparse coding model generalizes the prior distribution of the coefficients from a Gaussian noise vector to a sparse vector, the generator network retains the prior distribution of the coefficients as a Gaussian noise vector, but it generalizes the linear mapping from the coefficient vector to the signal in factor analysis to a non-linear mapping parametrized by a top-down convolutional neural network (or a so-called deconvolutional network).

However, unlike the sparse coding model, the generator network is a dense model without sparsity. In this paper, we try to fuse the sparse coding model and the generator network, or more specifically, we try to induce the sparse coding model from the generator network. To accomplish this, we use a simple mechanism where at each layer of the top-down generator network, we only select the top $k$ coefficients to be active and force all the other coefficients to be zeros. This leads to a top-down sparse coding generator network. The fusion leads to the following advantages. (1) The model can still be learned and inferred efficiently as in the original generator network. (2) The model can still generate realistic images. (3) The model incorporates the sparse coding principle and learns meaningful basis functions. Thus the proposed model naturally fuses the generator network and sparse coding model while maintaining their advantages.

Furthermore, the proposed model connects the sparsity principle to the compositionality principle. The model is explainable since visual patterns are hierarchical compositions of basis functions. In the language of AND-OR grammars (Zhu and Mumford 2007), the dictionary of the basis functions can be considered a large OR node, where each basis is a child node of this OR node and each generated image is itself an AND-OR structure, where AND means composition of the constituent basis functions and OR means different choices and configurations of the basis functions as well as variations of their coefficients.

* This work has been done when author working as visiting scholar in UCLA.

## 2 Generative Model with Sparse Activations

A typical generator has a top-down structure as illustrated in Figure 1. The input latent vector $Z \in \mathbb{R}^d$ is fed into a fully connected layer and through a Relu operation to produce featuremap1, or $fm^1 \in \mathbb{R}^{n_w^1 \times n_h^1 \times n_c^1}$, with non-linear activations. Then, featuremap1 is sent through the first deconvolutional layer with kernels $ker^1 \in \mathbb{R}^{n_c^1 \times h^1 \times h^1 \times n_c^2}$ and then through a Relu to produce featuremap2, $fm^2 \in \mathbb{R}^{n_w^2 \times n_h^2 \times n_c^2}$. The produced featuremap2 continues through deconvolutional layers with corresponding kernels $ker^i \in \mathbb{R}^{n_c^i \times h^i \times h^i \times n_c^{i+1}}$ after which the output image $Y \in \mathbb{R}^{n_w^o \times n_h^o \times 3}$ is generated with the non-linear sigmoid or tanh function. To introduce an explainable generative model, we apply the sparse operation on each layer's feature map such that the activations within the feature map are very sparse.

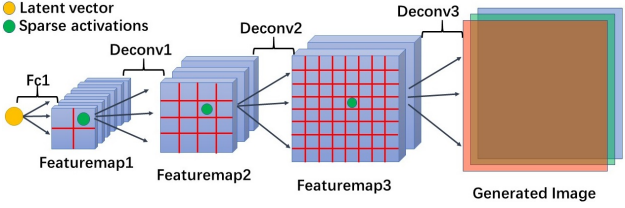

Figure 1: An illustration of the generator with sparse activations.

The sparse operation on the $i^{th}$ layer's feature map is defined as the $Top_K(\cdot, K)$ operation which selects and keeps the $K$ maximum elements within the tensor $fm^i$ and sets all other elements to zero, $fm_s^i = Top_K(fm^i, K)$.

### 2.1 Inducing And-Or-Graphs

After the sparse operation, most of the elements of the feature map become zeros, and very sparse activations survive. These sparse activations are located at different spatial directions and channels. (e.g.) For a feature map of size $n_w \times n_h \times n_c$ along the spatial directions, at most $n_w \times n_h$ activations survive. At a particular $(x_i, y_i)$ of these $n_w \times n_h$ locations, suppose $c_i$ activations survive along the channel direction, then we have, $K = \sum_{i=1}^{n_w \times n_h} c_i$, where $K$ is the number of overall sparse activations within the feature map. At each location $(x_i, y_i)$, we can define an 'AND' node, since a full image must consist of contents (activations) at all the spatial locations. For a particular spatial location $(x_i, y_i)$, we name the surviving activations along the channel direction as the 'OR' nodes, since at each location, we have different choices to utilize the kernels from the $c_i$ channels to construct the next layer's feature map or the final output image. Thus, these sparse activations form the And-Or Parsing Graph. The sparse activations are determined by the input latent vector and the learned parameters (weights and bias) of the generator. A different latent vector will generate a different parsing graph.

### 2.2 Inducing sparse coding

For the $i^{th}$ layer, the value of the $K^i$ sparse activations can be interpreted as the sparse coefficients, $s_j^i, j \in \{1, \ldots, K^i\}$. Suppose the corresponding $K$ basis functions are denoted as $H_j^i, j \in \{1, \ldots, K^i\}$, $H_j^i \in \mathbb{R}^{n_w^{i+1} \times n_h^{i+1} \times n_c^{i+1}}$, the sparse activations in the $(i+1)^{th}$ layer can be obtained as

$$\begin{aligned} fm_s^{i+1} &= ReLU(Top_K(fm^{i+1}, K^{i+1})) \\ &= ReLU(Top_K(\sum_{j=1}^{K^i} s_j^i \times H_j^i, K^{i+1})) \end{aligned} \quad (1)$$

It is worth to note that both $ReLU$ and the $Top_K$ operations are non-linear and can be seen as switches or masks that partitioning the space. However, after we obtain the parsing graph of a particular input latent vector $Z$, we can remember these masks of the chosen elements of both the $Top_K$ and $ReLU$ operations. Then, these non-linear operations will be changed to linear operations. Therefore, we can rewrite the sparse activations in the $(i+1)^{th}$ layer as

$$\begin{aligned} fm_s^{i+1} &= (fm^{i+1} \odot mask_T^{i+1}) \odot mask_R^{i+1} \\ &= (\sum_{j=1}^{K^i} s_j^i \times H_j^i \odot mask_T^{i+1}) \odot mask_R^{i+1} \end{aligned} \quad (2)$$

where $\odot$ denotes the dot product between two matrices. $mask_T$ and $mask_R$ are the masks of the $Top_K$ and the $ReLU$ results on the parsing graph, and both of them have the same size of $n_w^{i+1} \times n_h^{i+1} \times n_c^{i+1}$. The basis $H_j^i$ can be computed by setting the other sparse coefficients to be zeros, $s_k^i = 0, k \neq j$. Then we have

$$s_j^i \times H_j^i = fm_{s_j}^i \otimes ker^i + \frac{1}{K^i} bias^i \quad (3)$$

where $\otimes$ denotes the deconvolutional operation, $ker^i$ and $bias^i$ denote the kernels and bias at the $i^{th}$ layer, and $fm_{s_j}^i$ denotes the feature map at the $i^{th}$ layer with only one activation whose value is $s_j^i$. In other words, the feature map of each sparse activation at the $i^{th}$ layer, can be written as,

$$\begin{aligned} fm_{s_j}^i &= (fm^i \odot mask_T^{i,j}) \odot mask_R^i \\ &= (\sum_{j=1}^{K^{i-1}} s_j^{i-1} \times H_j^{i-1} \odot mask_T^{i,j}) \odot mask_R^i \end{aligned} \quad (4)$$

where $mask_T^{i,j}$ denotes the mask that only chooses the $j^{th}$ largest elements from the top $K^i$ activations at layer i.

Moreover, the final generated image $Y$ can also be presented in the unified sparse coding framework as

$$Y = tanh(\sum_{j=1}^{K^i} s_j^i B_j^i) \quad (5)$$

where $B_j^i \in \mathbb{R}^{n_w^o \times n_h^o \times 3}$ is the synthesis basis corresponding to the $j^{th}$ sparse activation in the $i^{th}$ layer, and can be

computed as

$$\begin{aligned} B_j^i &= fm^L \odot mask_T^L \\ &= fm_s^{L-1} \otimes ker^{L-1} + \frac{1}{K^{L-1}} bias^{L-1} \end{aligned} \quad (6)$$

where $L$ is number of the bottom layer, and $fm_s^{L-1}$ can be computed by Eq. (2) and (3) with recursion.

### 2.3 Learning and Inference

The generator model with sparse activations can be expressed as

$$\begin{aligned} Z &\sim \mathrm{N}(0, I_d), \\ Y &= g(Z; \theta, t_k) + \epsilon, \ \epsilon \sim \mathrm{N}(0, \sigma^2 I_D). \end{aligned} \quad (7)$$

$g(Z; \theta, t_k)$ is a top-down sparse ConvNet defined by both the parameters $\theta$ which includes the weight and bias parameters and $t_k$ which includes the number of each layer's maximum surviving activations. The latent vector $Z$ is mapped to the signal $Y$ by the sparse ConvNet $g$. To learn this generator model, we introduce a learning and inference algorithm, without designing and learning extra inference networks. Specifically, the proposed model can be trained by maximizing the log-likelihood on the training dataset $\{Y_i, i = 1, \ldots, N\}$,

$$\begin{aligned} L(\theta) &= \frac{1}{N} \sum_{i=1}^{N} \log P(Y_i; \theta, t_k) \\ &= \frac{1}{N} \sum_{i=1}^{N} \log \int P(Y_i, Z; \theta, t_k) dZ. \end{aligned} \quad (8)$$

The uncertainty in inferring $Z$ is taken into account by the above observed-data log-likelihood. We can compute $\theta$ by minimizing the Kullback-Leibler divergence $\mathrm{KL}(P_{\mathrm{data}}|P_\theta)$ from the data distribution $P_{\mathrm{data}}$ to the model distribution $P_\theta$.

The gradient of $L(\theta)$ is obtained according to the following equation which is related to the EM algorithm

$$\begin{aligned} &\frac{\partial}{\partial \theta} \log P(Y; \theta, t_k) \\ &= \frac{1}{P(Y; \theta, t_k)} \frac{\partial}{\partial \theta} \int P(Y, Z; \theta, t_k) dZ \\ &= \mathrm{E}_{P(Z|Y;\theta,t_k)} \left[ \frac{\partial}{\partial \theta} \log P(Y, Z; \theta, t_k) \right]. \end{aligned} \quad (9)$$

In general, the expectation in (9) is analytically intractable, and needs to be approximated by an MCMC method that samples from the posterior $P(Z|Y; \theta, t_k) \propto p(Y, Z; \theta, t_k)$, such as the Langevin dynamic inference, which iterates

$$Z_{\tau+1} = Z_\tau + \frac{\delta^2}{2} \frac{\partial}{\partial Z} \log P(Z_\tau, Y; \theta, t_k) + \delta \mathcal{E}_\tau, \quad (10)$$

where $\tau$ indexes the time step, $\delta$ is the step size, and $\mathcal{E}_\tau$ denotes the noise term, $\mathcal{E}_\tau \sim \mathrm{N}(0, I_d)$. The log of the joint density in Eq.(10) can be evaluated by

$$\begin{aligned} &\log p(Y, Z; \theta, t_k) = \log [p(Z) p(Y|Z; \theta, t_k)] \\ &= -\frac{1}{2\sigma^2} \|Y - g(Z; \theta, t_k)\|^2 - \frac{1}{2} \|Z\|^2 + C \end{aligned} \quad (11)$$

where $C$ is the constant term, and is independent of $Z$ and $Y$. It can be shown that, given sufficient transition steps, the $Z$ obtained from this procedure follows the joint posterior distribution. For each training example $Y_i$, we run the Langevin dynamics in Eq.(10) to get the corresponding posterior sample $Z$. The sample is then used for gradient computation in Eq.(9). More precisely, the parameter $\theta$ is learned through Monte Carlo approximation:

$$\begin{aligned} \frac{\partial}{\partial \theta} L(\theta) &\approx \frac{1}{N} \sum_{i=1}^{N} \frac{\partial}{\partial \theta} \log p(Y_i, Z_i; \theta, t_k) \\ &= \frac{1}{N} \sum_{i=1}^{N} \frac{1}{\sigma^2} (Y_i - g(Z_i; \theta, t_k)) \frac{\partial}{\partial \theta} g(Z_i; \theta, t_k). \end{aligned} \quad (12)$$

### 2.4 Extension with Energy Based Model

It is well known that using squared Euclidean distance alone to train generator networks often yields blurry reconstruction results, since the precise location information of details may not be preserved, and the $L_2$ loss in the image space leads to averaging all likely locations. In order to improve the generative performance, we employ a descriptor to describe the distribution and the context information to help the generator produce better results. A descriptor model is also a kind of energy-based model, and is in the form of exponential tilting of a reference distribution

$$P(Y; \phi) = \frac{1}{Z(\phi)} \exp\left[f(Y; \phi)\right] q(Y). \quad (13)$$

$q(Y)$ is the reference distribution such as Gaussian white noise. $f(Y; \phi)$ is a bottom-up ConvNet which maps the image $Y$ to the feature statistics. $f(Y; \phi)$ is also known as the energy function. $Z(\phi) = \int \exp\left[f(Y; \phi)\right] q(Y) dY = \mathrm{E}_q\{\exp[f(Y; \phi)]\}$ is the normalizing constant. The descriptor can capture the distribution and context of the training images by maximizing the log-likelihood function. To improve the generator's performance, we feed $\tilde{Y}_i$ which is sampled from the descriptor into Eq. (12) to replace $Y_i$. Since $\tilde{Y}_i$ contains meaningful distribution and context information of the training data learned from the energy-based model, it is easier for the generator to learn from $\tilde{Y}_i$ than directly learn from $Y_i$. Learning from $Y_i$ directly is difficult and the generated image may look blurry, since it requires per pixel reconstruction. The model can also be learned by VAE and GAN, except that VAE requires an extra inference model and does not have strong synthesis power, while GAN cannot do inference and has mode collapsing issue.

## 3 Experimental Results

To demonstrate the meaningful hierarchical representation power of the proposed model, we show the hierarchical representation results of the generator. The generator network contains one fully-connected layer and two convolutional layers. We train the sparse generator on 5k images from CelebA benchmark dataset.

Figure 2 shows the learned kernels of the sparse generator with two convolutional layers. When the visual fields

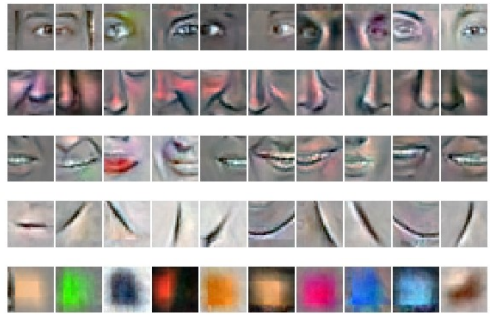

(a) Convolutional kernels in the second (bottom) layer.

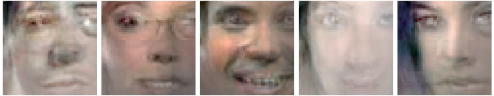

(b) Hierarchical representation of the kernels in the first layer.

Figure 2: Hierarchical representation of the kernels learned by the sparse generator with two convolutional layers in the (a) bottom layer (b) top layer.

of the bottom layer is relatively large, the bottom convolutional kernels can learn different facial parts, such as eyes, noses, mouths, chins and nasolabial folds, which are shown in Figure 2 (a). As we can observe from Figure 2 (b), the top layer's kernels have learned some trends to combine the facial parts in the bottom layer's kernels into a whole face. It is worth to note that without the proposed sparse-K operation on the activations (or feature maps), we cannot obtain these meaningful results in both layer's kernel. With the sparse-K operation, the energies are forced to be collected into very few activation, which makes the corresponding kernels' content meaningful.

To understand the sparse generator's working principle, we further explain the images' generating process as a hierarchical And-Or Parsing Graph as shown in Figure 3. Specifically, as we can observe from the left part of Figure 3, at the bottom convolutional layer, the basis functions are the convolutional kernels themselves. A generated image is partitioned by $4 \times 4 = 16$ 'AND' nodes, and each 'AND' node contains several 'OR' nodes along the channel direction. The 'AND' node at each location (e.g. eyes, nose, or mouth) of the face contains the 'OR' nodes consisting of the corresponding colors and shapes from the same category. For different faces, the configuration of the 'OR' nodes are different, although they share the same dictionaries of bases. Due to these different configurations, we can generate different kinds of facial parts and different faces.

The generated face can also be described by the coefficients of the sparse activations from any layer multiplied by the corresponding basis functions according to Eq. (5). Specifically, as we can observe from the right part of Figure 3, at the top layer, a face image is partitioned into $2 \times 2 = 4$ 'AND' nodes, and each 'AND' node contains several 'OR' nodes. These 'OR' nodes (the basis functions at the top layer) reveal high-level semantics, and are the combinations of the basis functions at the bottom later.

The proposed model can learn both other objects and tex-ture patterns. Figure 4 shows some basis functions from the bottom layer of the generator with two convolutional layers.

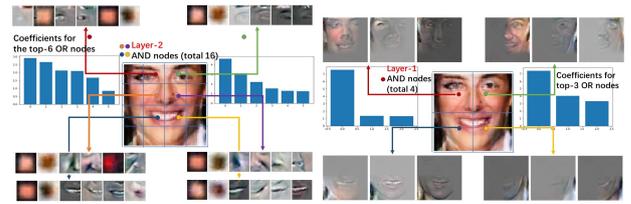

Figure 3: Analyzing generated face of sparse generator with two convolutional layers using AND-OR Parsing Graph.

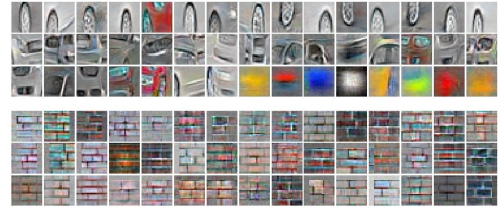

Figure 4: Learned bottom kernels from generator with two convolutional layers on the cars and brickwall images.

## 4   Conclusion

In this study, we introduce a sparse generative model which naturally combines the generator network and sparse coding model while maintaining their advantages. Our experiments show that the model can learn meaningful dictionaries at different layers.